\documentclass{article} 
\usepackage[preprint]{colm2026_conference}

\usepackage{microtype}
\usepackage{hyperref}
\usepackage{url}
\usepackage{booktabs}
\usepackage{graphicx}
\usepackage{tikz}
\usetikzlibrary{positioning, shapes.multipart, arrows.meta}
\usepackage{makecell}
\usepackage{array}
\definecolor{matchorange}{RGB}{230,120,20} 
\usepackage{adjustbox} 
\usepackage[table]{xcolor}
\definecolor{matchorange}{RGB}{230,120,20} 

\usepackage{times}
\usepackage{latexsym}
\usepackage{amsmath}
\usepackage{amssymb}
\usepackage{CJKutf8}
\usepackage{float}
\usepackage{subcaption}  

\usepackage{lineno}

\definecolor{darkblue}{rgb}{0, 0, 0.5}
\hypersetup{colorlinks=true, citecolor=darkblue, linkcolor=darkblue, urlcolor=darkblue}

\title{MindAlign: Decoding Inner Speech from fMRI Signals via Multimodal Embedding Alignment under Limited Data}

\author{%
Muxuan Liu \& Ichiro Kobayashi\\
Graduate School of Humanities and Sciences\\
Ochanomizu University\\
\texttt{\{liu.muxuan,koba\}@is.ocha.ac.jp}\\
\And
Satoshi Nishida\\
Center for Information and Neural Networks\\
Advanced ICT Research Institute\\
National Institute of Information and \\Communications Technology\\
\texttt{s-nishida@nict.go.jp}
}

\begin{document}

\ifcolmsubmission
\linenumbers
\fi

\maketitle

\begin{abstract}
Decoding inner speech from non-invasive brain signals remains a fundamental challenge due to the absence of overt linguistic output, limited training data, and large inter-subject variability. Existing brain-to-text approaches often rely on task-specific decoder fine-tuning, which restricts scalability and complicates adaptation to new participants.
We propose \textbf{MindAlign}, a decoupled two-stage brain-to-language framework that enables open-ended text generation from fMRI signals without modifying the underlying language model. The first stage learns a subject-specific neural-semantic alignment that maps fMRI activity into a shared multimodal semantic space, extracting a latent semantic sketch of the internally generated sentence. The second stage integrates this sketch with visual context to prompt a frozen multimodal language model for free-form generation.
Experiments on fMRI data collected during silent image description demonstrate that the proposed approach consistently outperforms fMRI-only and random baselines. We further show that the learned semantic-to-language projection can generalize across subjects, enabling effective decoding when paired with subject-specific neural alignment. These results indicate that neural signals modulate semantic content beyond image-driven priors, supporting a scalable and modular direction for brain-to-text decoding.
\end{abstract}

\section{Introduction}
Translating internal speech directly from brain activity into natural language is a long-standing goal in both neuroscience and artificial intelligence. Functional magnetic resonance imaging (fMRI) is non-invasive and offers high spatial resolution, yet its low temporal resolution makes open-ended language decoding particularly challenging.

Most prior brain-to-text research relies on tightly time-aligned stimuli such as spoken stories or movie subtitles \citep{pereira2018toward, chen2024openvocabulary}. Others constrain outputs to predefined vocabularies \citep{yargholi2016brain, zou2022cross}, limiting free-form decoding of internally generated thoughts.

We propose \textbf{MindAlign}, a two-stage framework for decoding free inner speech in Japanese from fMRI signals collected during natural image viewing. Rather than aligning fMRI with word-level timing, we train an encoder to produce \emph{semantic sketches}, which are fed into a frozen vision-language decoder through soft prefix tuning, optionally incorporating image features.

\textbf{Contributions.}  

(1) We plan to release a new fMRI dataset that pairs image-evoked inner speech with voxel time-series from five participants. 

(2) We propose a two-stage decoder that works \emph{without external audio--token alignment}, and conduct both subject-wise and cross-subject evaluations.

(3) With approximately 2,000 samples from five participants, the model generates fluent Japanese sentences and consistently outperforms image-only and random baselines across multiple metrics, demonstrating free-form inner-speech decoding in Japanese from fMRI.

\section{Related Work}
\begin{table}[ht]
\centering
\scriptsize
\renewcommand{\arraystretch}{1.1}
\begin{tabular}{|p{3.5cm}|p{7cm}|p{1.5cm}|}
\hline
\textbf{Model / Paper} & \textbf{Training Modality (A $\rightarrow$ B: predict B from A)} & \textbf{Subject Setting} \\
\hline
\citet{pereira2018toward} & fMRI (word reading with disambiguation contexts) $\rightarrow$ word-level semantic vector (from GloVe embedding space) & subject-wise \\
\hline
MindEye~\citep{scotti2023reconstructing} & fMRI (image viewing) $\rightarrow$ image (retrieved or reconstructed via CLIP embedding) & subject-wise \\
\hline
MindEye2~\citep{scotti2024mindeye} & fMRI (image viewing) $\rightarrow$ image (via CLIP embedding; shared-subject training then adapt to target subject) & subject-agnostic \\
\hline
MindBridge~\citep{wang2024mindbridge} & fMRI (image viewing) $\rightarrow$ image (reconstructed from CLIP embedding) & subject-agnostic \\
\hline
\citet{tang2023semantic} & fMRI (auditory/imagined story) $\rightarrow$ text (story) & subject-wise \\
\hline
Thought2Text~\citep{Mishra2025} & EEG (image viewing) + image $\rightarrow$ text (pseudo-labels from LLMs; decoded by finetuned model) & subject-wise \\
\hline
BP-GPT~\citep{chen2024openvocabulary} & fMRI (auditory story) $\rightarrow$ text (story) & subject-wise \\
\hline
BrainChat~\citep{huang2024brainchat} & fMRI (image viewing) + image + text $\rightarrow$ text (caption/QA from external datasets, not subject-generated) & subject-wise \\
\hline
MindFormer~\citep{han2024mindformer} & fMRI (image viewing) $\rightarrow$ image / text (via semantic image features and LLM) & subject-agnostic \\
\hline
BrainCLIP~\citep{Ma2025} & fMRI (image viewing) $\rightarrow$ image/text embedding (CLIP supervision) & subject-wise \\
\hline
BrainLLM~\citep{ye2025generative} & fMRI (visual/auditory language stimuli) $\rightarrow$ text (stimulus sentence continuation) & subject-wise \\
\hline
MindLLM~\citep{qiu2025mindllm} & fMRI (image viewing) $\rightarrow$ text (caption/QA/reasoning; instruction-tuned LLM decoding) & subject-agnostic \\
\hline
LaVCa~\citep{matsuyama2025lavca} & fMRI (visual cortex responses) $\rightarrow$ text (LLM-assisted captions for voxel selectivity / image semantics) & subject-wise \\
\hline
Interpretable fMRI Captioning~\citep{shen2025interpretable} & fMRI (image viewing) $\rightarrow$ text (caption; shared embedding contrastive learning) & subject-wise \\
\hline
Efficient Multi Subject Visual Reconstruction~\citep{zangos2025efficient} & fMRI (image viewing) $\rightarrow$ image (aligned common representation for cross-subject reconstruction) & subject-agnostic \\
\hline
Mind Captioning~\citep{horikawa2025mind} & fMRI (visual experience and recall) $\rightarrow$ text (descriptive captions generated by language models)  & subject-agnostic\\
\hline
\textbf{MindAlign (Ours)} & fMRI (image-evoked inner speech) + image $\rightarrow$ text (subject-generated inner speech) & subject-wise\\
\hline
\end{tabular}
\caption{Overview of brain decoding models. A model is \emph{subject-wise} if trained/tested per participant, and \emph{subject-agnostic} if shared across participants.}
\label{tab:brain_data_minimal}
\end{table}

Table~\ref{tab:brain_data_minimal} overviews major research routes of brain‑signal decoding.
Early studies mapped \emph{static} fMRI responses to \emph{word‑level semantic vectors} \citep{pereira2018toward}, but were limited to small lexicons and per‑subject tuning.  
Image‑focused methods such as MindEye and its cross‑subject variants \citep{scotti2023reconstructing,scotti2024mindeye,wang2024mindbridge,Ma2025} produce impressive visual reconstructions, yet they are trained only on the Natural Scenes Dataset (NSD)~\citep{allen2022massive} and flatten high‑resolution 3‑D voxel maps, losing cortical geometry.

Sentence‑level models \citep{tang2023semantic,chen2024openvocabulary,ye2025generative} aim to decode \emph{continuous text}.  \citet{tang2023semantic} first let GPT generate candidate sentences and then apply ridge regression to map those sentence embeddings back into voxel space, comparing the result with measured fMRI; this pipeline demands frame‑accurate alignment among audio, tokens, and TRs and inherits the combined errors of the language model and the regressor.  
BP‑GPT~\citep{chen2024openvocabulary} faces the same temporal‑mismatch problem: each fMRI volume spans several words, and its “GPT‑2 generation $\rightarrow$ ridge back‑projection” design compounds alignment errors, preventing reliable word‑level decoding.  
BrainLLM~\citep{ye2025generative} pursues open‑vocabulary generation, but its PCA compression (50k → 1,000 voxels) and frozen linear map into a pretrained LLM embedding space ignore cortical topology and temporal dynamics, leading to syntactic and semantic inaccuracies in the decoded sentences.

Recent multimodal systems pair neural signals with captions or VQA answers from public corpora \citep{qiu2025mindllm,huang2024brainchat}—these approaches usually rely on image–text datasets as pseudo‑labels and thus lack participant‑generated language—or switch to EEG with coarse spatial detail \citep{Mishra2025}; MindFormer~\citep{han2024mindformer} adds diffusion priors for cross‑subject images.  Related captioning-oriented studies further explore fMRI-to-text mappings by generating descriptive language for visual content. 
LaVCa~\citep{matsuyama2025lavca} employs large language models to produce captions that explain voxel selectivity in the visual cortex, emphasizing interpretability rather than sentence reconstruction. 
Similarly, Interpretable fMRI Captioning~\citep{shen2025interpretable} formulates stimulus image captioning as a shared-embedding contrastive learning problem. 
Most recently, Mind Captioning~\citep{horikawa2025mind} demonstrates that descriptive text can be generated from fMRI signals associated with visual experience and recall, but it focuses on generic scene descriptions rather than recovering subject-generated internal language.

Two key gaps remain:  
(i) most studies rely on externally provided captions rather than participants’ own inner speech, and  
(ii) existing decoding methods struggle to preserve subject-specific semantics from fMRI signals.  
We introduce a novel fMRI dataset of image-evoked inner speech and a two-stage \textbf{MindAlign} framework that maps neural activity to input embeddings.

\begin{CJK}{UTF8}{min}
\begin{figure}[h]
    \centering
    \includegraphics[width=1\textwidth]{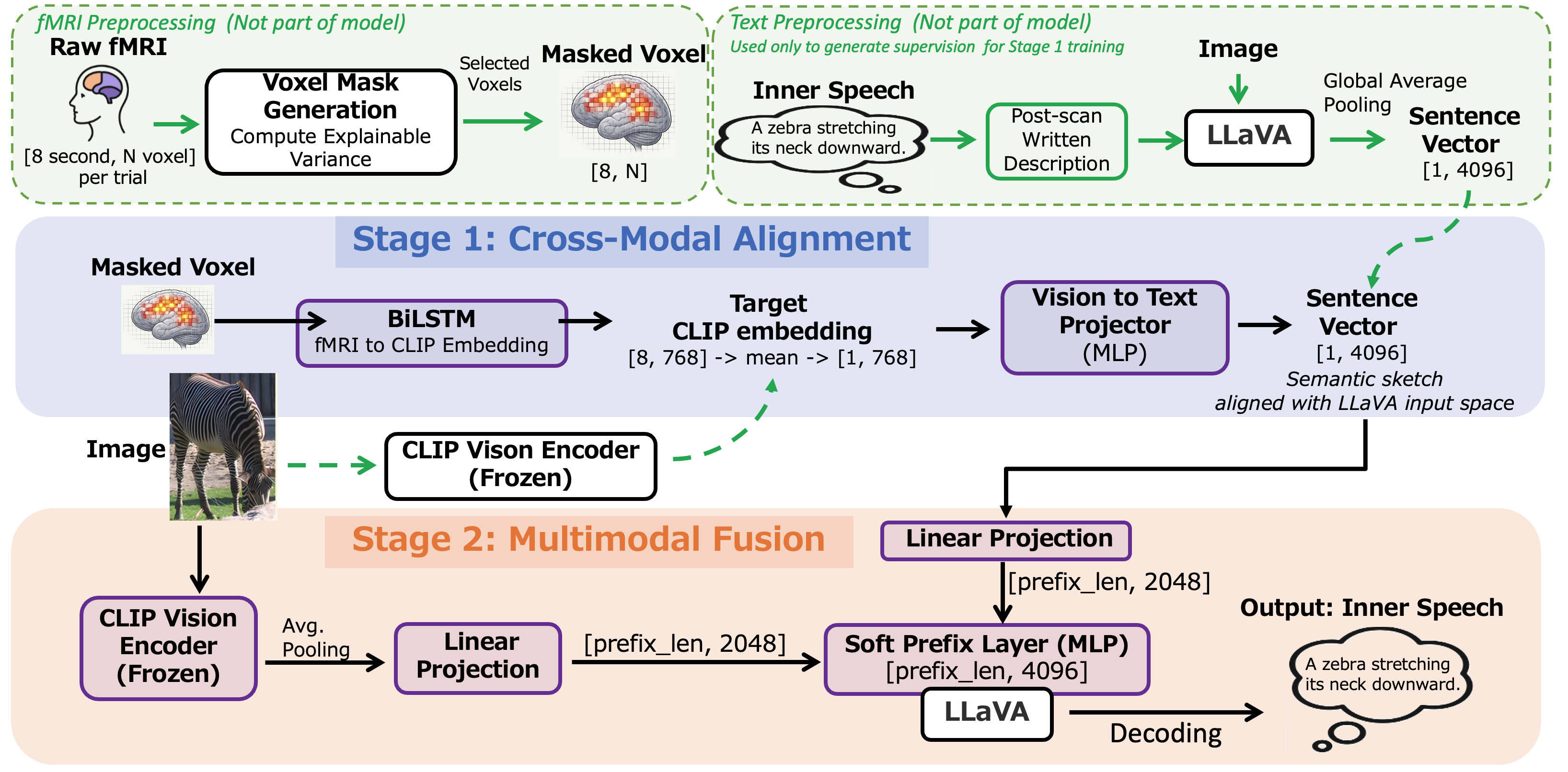}
    \caption{
Overview of MindAlign, our proposed two-stage brain-to-language decoding framework. CLIP embeddings are used only as an intermediate semantic anchor during Stage-1 training; the final aligned representation is expressed in the LLaVA input embedding space and passed to Stage-2.\\
}
\label{fig:framework}
\end{figure}
\end{CJK}

\color{black}
\section{Method}
\label{sec:method}

We propose a two-stage brain-to-language decoding framework that reconstructs internally imagined speech from non-invasive fMRI recordings. The framework consists of (1) \textit{Cross-Modal Alignment}, which maps fMRI signals into a pretrained multimodal semantic space, and (2) \textit{Multimodal Fusion Generative Decoding}, which uses the aligned representations to prompt a Large Language Model (LLM) for text generation.

\subsection{Data Preprocessing}
To reduce dimensionality and suppress noise, the fMRI inputs to the encoder are restricted to a subset of voxels selected based on explainable variance (EV).
Voxel selection is performed at the dataset preprocessing stage using category-based EV filtering, and the resulting voxel mask is fixed and shared across all splits.
Details are provided in Section~\ref{sec:dataset}.


\subsection{Stage 1: Cross-Modal Alignment}

The objective of Stage 1 is to extract a latent semantic representation from high-dimensional, noisy fMRI data. During training, this representation is constrained using CLIP~\cite{radford2021learning} visual embeddings as an intermediate semantic supervision signal, while the final output is projected into the LLaVA input embedding space for downstream decoding.

\paragraph{Target Feature Extraction}
For each stimulus image $I$, we extract its visual embedding $\mathbf{z}_{vis} \in \mathbb{R}^{d_{clip}}$ using a pretrained \texttt{CLIP-ViT-L/14} model \footnote{\url{https://huggingface.co/openai/clip-vit-large-patch14}}. These embeddings are $\ell_2$-normalized and serve as an intermediate semantic supervision signal during Stage-1 training, rather than the final representation space. Additionally, we extract embeddings from the input layer of \texttt{LLaVA-1.6-vicuna-7B-hf} and apply Global Average Pooling (GAP) to derive a fixed-dimensional sentence-level vector $\mathbf{z}_{lang}$, providing auxiliary linguistic supervision.

\paragraph{fMRI Encoder Architecture}
The encoder employs a Bidirectional LSTM (Bi-LSTM)~\cite{huang2015bidirectional} to capture the temporal dynamics of the fMRI signal ($T=8$ TRs). The process involves:
\begin{enumerate}
    \item \textbf{Feature Projection:} A linear layer projects voxel values into a feature space: $\mathbf{h}_t = \text{ReLU}(\text{LN}(\mathbf{W}_{in} \mathbf{x}_t + \mathbf{b}_{in}))$.
    \item \textbf{Temporal Encoding:} A multi-layer Bi-LSTM aggregates information across the sequence: $\mathbf{o}_t = \text{Bi-LSTM}(\mathbf{h}_t)$.
    \item \textbf{Global Pooling:} We apply GAP over the temporal dimension to obtain a single sentence-level representation: $\mathbf{o}_{avg} = \frac{1}{T} \sum_{t=1}^T \mathbf{o}_t$.
    \item \textbf{Semantic Mapping:} An MLP aligns $\mathbf{o}_{avg}$ with the CLIP embedding space ($d_{clip}=768$) for semantic supervision, followed by a linear projection into the LLaVA input embedding space ($d_{llm}=4096$).
\end{enumerate}

\paragraph{Training Objectives and Augmentation}
We optimize the encoder using a hybrid loss function:
\begin{equation}
\begin{aligned}
\mathcal{L}_{\text{total}} =\;
& \lambda_{\text{nce}} \mathcal{L}_{\text{NCE}}
+ \lambda_{\text{cos}} \mathcal{L}_{\text{cos}}
+ \lambda_{\text{mse}} \mathcal{L}_{\text{MSE}}
+ \lambda_{\text{cls}} \mathcal{L}_{\text{cls}}
\end{aligned}
\end{equation}
where $\mathcal{L}_{NCE}$ is the InfoNCE contrastive loss, $\mathcal{L}_{cos}$ and $\mathcal{L}_{MSE}$ provide point-wise supervision, and $\mathcal{L}_{cls}$ is an auxiliary cross-entropy loss using COCO category labels. To enhance robustness, we incorporate \textit{Intra-category Mixup}, a category-constrained variant of Mixup~\citep{zhang2017mixup}, which has been shown to improve performance and reduce overfitting in fMRI-based deep learning models~\citep{smucny2022data}. Additionally, we apply spatially constrained noise injection as a form of regularization~\citep{shorten2019survey}, which perturbs voxel activations within localized regions to improve robustness to input noise.

\subsection{Stage-2: Multimodal Fusion Generative Decoding}

As illustrated in the lower part of Appendix Figure~\ref{fig:method-overview} and summarized in the Stage 2 panel of Figure~\ref{fig:framework}, Stage 2 conditions language generation on fMRI inputs via multimodal soft prefix tuning, leveraging the frozen \texttt{LLaVA-1.6-vicuna-7B-hf} model.

\paragraph{Multimodal Prefix Generation}
The objective of Stage 2 is to construct a soft prefix $\mathbf{P} \in \mathbb{R}^{L \times d_{llm}}$ (where $L$ is a hyperparameter representing the prefix length, set to 8 in our experiments) that guides the LLM. The generation process involves a dual-stream projection and fusion mechanism:
\begin{enumerate}
    \item \textbf{Neural Stream:} The frozen Stage~1 encoder outputs a sentence-level semantic vector $\mathbf{e}_{fmri} \in \mathbb{R}^{4096}$ in the LLaVA input embedding space. This vector is then linearly projected into a neural prefix component of shape $[B, L \times (d_{llm}/2)]$ via a linear layer $\mathbf{W}_{f}$, where $d_{llm}/2 = 2048$.
    \item \textbf{Visual Stream:} Simultaneously, visual features $\mathbf{v} \in \mathbb{R}^{d_{v}}$ are extracted from the stimulus image via a frozen CLIP ViT-L/14 encoder. These features are projected into a visual prefix component of shape $[B, L \times (d_{llm}/2)]$ via a linear layer $\mathbf{W}_{v}$.
    \item \textbf{Feature Fusion:} The neural and visual prefix components are concatenated along the feature dimension to form a combined representation of shape $[B, L \times d_{llm}]$. This representation is then reshaped to $[B, L, d_{llm}]$ and passed through a fusion MLP (comprising Linear, LayerNorm, and GELU layers) to produce the final soft prefix $\mathbf{P}$. In this setup, the initial projection to $d_{llm}/2$ (2048) ensures that both modalities contribute equally to the fused prefix space before the final non-linear integration.
\end{enumerate}

While LLaVA typically uses CLIP-derived image embeddings to construct visual prefixes, we replace this mechanism with a joint mapping from fMRI and image features, allowing our model to condition generation on brain signals. This prefix is prepended to the BOS token and passed to the frozen LLaVA decoder via the input embedding interface, without modifying any decoder or backbone parameters. Only the projection and fusion modules are updated during training. We fix the prefix length to 8 based on the default experimental configuration, with additional variants discussed in Appendix~\ref{appendix:prefix_length}.

\paragraph{Generative Training}
The model is trained using teacher forcing with cross-entropy loss, minimizing the negative log-likelihood of the ground-truth token sequence $y_{1:T}$ given the soft prefix $\mathbf{P}$:
\begin{equation}
\mathcal{L}_{gen} = - \sum_{i=1}^{N} \log P(y_i \mid y_{<i}, \mathbf{P})
\label{eq:stage2-loss}
\end{equation}
To prevent the model from predicting tokens in the prefix region, the corresponding positions in the target label sequence are masked using a label value of $-100$, following the standard convention in sequence modeling.

\section{Experiments}

\subsection{Dataset}
\label{sec:dataset}

We evaluate our method using an fMRI-image-text dataset collected during a naturalistic inner speech task.
Five participants (LD0001--LD0006; LD0003 withdrew during the experiment) viewed images from the COCO dataset~\cite{lin2014microsoft} and silently generated Japanese descriptions during fMRI scanning, without overt speech or lip movement.
All descriptions were produced in Japanese, consistent with the participants’ native language.

As shown in Appendix Figure~\ref{fig:stimulus_example}, each trial consisted of four consecutive 8-second segments.
In the first segment, the participant described the entire image.
In the remaining three segments, the participant described individual objects highlighted by white bounding boxes, with the instruction to describe the largest object when multiple objects were present.
Functional images were acquired with a repetition time (TR) of 1 second. The response signals of the voxels obtained for each subject are referenced in the Appendix Table~\ref{tab:participants}.

The experimental paradigm comprised five blocks of 16 trials per scanning day, yielding 80 trials per day and 560 image trials across seven scanning days. Each image trial included four consecutive description tasks, with the corresponding fMRI measurements segmented into 36-TR samples. A subset of images was repeated for validation.
After scanning, the participant provided one written description per image, which was used as the semantic target for all four associated fMRI segments.
Additional details on participant demographics, task training, and label validation are provided in Appendix~\ref{appendix:participant_info}.

\paragraph{Semantic Supervision Targets.}
To construct supervision targets for Stage~1, we extract visual embeddings from each stimulus image using the pretrained \texttt{CLIP-ViT-L/14} model.
In addition, we extract semantic sketch in the LLM embedding space from the input embedding layer of \texttt{LLaVA-1.6-vicuna-7B-hf}.
Since inner speech descriptions vary in length, we apply global average pooling over the token dimension to obtain fixed-dimensional sentence-level representations.
These pooled linguistic embeddings, together with CLIP visual embeddings, serve as semantic targets for aligning fMRI-derived features in a unified multimodal space.

\paragraph{ROI Selection and Voxel Filtering.}
To reduce input dimensionality and improve signal reliability, we apply a category-based voxel selection procedure based on explainable variance (EV)~\cite{Sahani2002, Hsu2004, Schoppe2016}, which quantifies the proportion of response variance that is consistently driven by stimulus category rather than trial-level noise.
For each semantic category, voxels with the highest EV values are selected, and a global union mask is constructed across all categories.
This procedure allows the model to focus on voxels exhibiting reliable, category-relevant responses while substantially reducing noise-dominated dimensions.
The formal definition of EV and the detailed selection procedure are described in Appendix~\ref{appendix:ev}.


\subsection{Subject-wise Evaluation Protocol}
\label{sec:subject-wise-protocol}

Due to substantial inter-subject variability in neural responses, we adopt a subject-wise training and evaluation protocol as our primary experimental setting. For each participant, a dedicated Stage-1 encoder is trained to map that subject’s fMRI signals into a shared semantic space guided by CLIP.
The Stage-2 prefix projector is then trained using the CLIP-aligned features produced by the corresponding subject-specific Stage-1 encoder.
All training, validation, and test splits are constructed independently for each subject, and no neural data are shared across subjects during training.
\color{black}
\subsection{Ablation Study: Contribution of fMRI and Image Modalities}
\label{sec:ablation}

To assess the contribution of different input modalities, we compare three variants of the decoding pipeline (see Appendix~\ref{sec:training-protocol} for protocol details):

\begin{itemize}
    \item \textbf{fMRI-only (Ours)}: A soft prefix is generated solely from brain signals by projecting the Stage-1 fMRI semantic representation into a $[\text{prefix\_len}, 4096]$ decoder prompt.
    \item \textbf{fMRI + Image (Ours)}: fMRI semantics and CLIP image features are independently projected to $[\text{prefix\_len}, 2048]$ and concatenated into a $[\text{prefix\_len}, 4096]$ multimodal prefix, followed by a lightweight fusion MLP.    
    \item \textbf{Random fMRI + Image}: As a control, fMRI inputs are replaced with Gaussian noise at inference to test whether brain-derived signals add information beyond images alone.
\end{itemize}

Both fMRI-only and fMRI+Image models share the same decoder (LLaVA) and adopt the same prefix-tuning interface using input embeddings, without modifying decoder weights. During training, only the prefix projection (and fusion, if applicable) layers are updated.



\begin{table*}[t]
\centering
\scriptsize
\renewcommand{\arraystretch}{1.15}
\setlength{\tabcolsep}{4pt}

\begin{subtable}[t]{\textwidth}
\centering
\caption*{\textbf{(a)} Subject-wise comparison: fMRI-only vs fMRI+Image.}
\begin{adjustbox}{max width=\linewidth}
\begin{tabular}{
l
cc
cc
cc
cc
cc
}
\toprule
 & \multicolumn{10}{c}{\textbf{Subject}} \\
\cmidrule(lr){2-11}
\textbf{Metric}
& \multicolumn{2}{c}{LD0001}
& \multicolumn{2}{c}{LD0002}
& \multicolumn{2}{c}{LD0004}
& \multicolumn{2}{c}{LD0005}
& \multicolumn{2}{c}{LD0006} \\
\cmidrule(lr){2-3}\cmidrule(lr){4-5}\cmidrule(lr){6-7}\cmidrule(lr){8-9}\cmidrule(lr){10-11}
& fMRI & fMRI+Img
& fMRI & fMRI+Img
& fMRI & fMRI+Img
& fMRI & fMRI+Img
& fMRI & fMRI+Img \\
\midrule
TF-IDF Cosine~\citep{salton1988term}
& 0.0844 & \textbf{0.1100}
& 0.1172 & \textbf{0.1183}
& 0.0741 & \textbf{0.1122}
& 0.0858 & \textbf{0.1522}
& 0.1383 & \textbf{0.1499} \\
ChrF~\citep{popovic-2015-chrf}
& 0.1193 & \textbf{0.1451}
& \textbf{0.1530} & 0.1481
& 0.1150 & \textbf{0.1443}
& 0.1451 & \textbf{0.2191}
& 0.1879 & \textbf{0.1984} \\
Levenshtein Similarity~\citep{levenshtein1966binary}
& 0.1489 & \textbf{0.1988}
& 0.2135 & \textbf{0.2194}
& 0.1568 & \textbf{0.2063}
& 0.2171 & \textbf{0.2803}
& 0.2780 & \textbf{0.3151} \\
BLEU~\citep{papineni-etal-2002-bleu} 
& 0.0866 & \textbf{0.1262}
& 0.1324 & \textbf{0.1353}
& 0.0849 & \textbf{0.1284}
& 0.1274 & \textbf{0.1951}
& 0.1712 & \textbf{0.1916} \\
ROUGE-L~\citep{lin-2004-rouge}
& 0.2512 & \textbf{0.3039}
& 0.3354 & \textbf{0.3465}
& 0.2606 & \textbf{0.3278}
& 0.3230 & \textbf{0.3949}
& 0.3910 & \textbf{0.4222} \\
BERT-Score~\citep{zhang2020bertscoreevaluatingtextgeneration} 
& 0.7215 & \textbf{0.7384}
& 0.7493 & \textbf{0.7584}
& 0.7159 & \textbf{0.7416}
& 0.7258 & \textbf{0.7562}
& 0.7683 & \textbf{0.7800} \\
\bottomrule
\end{tabular}
\end{adjustbox}
\end{subtable}

\vspace{4pt}

\begin{subtable}[t]{\textwidth}
\centering
\caption*{\textbf{(b)} Baseline comparison: Original LLaVA (image-only) vs Random fMRI+Image. 
\;(\textit{Note: Original LLaVA scores are identical across subjects.})}
\begin{adjustbox}{max width=\linewidth}
\begin{tabular}{l c c c c c c}
\toprule
\textbf{Metric}
& \textbf{LLaVA} 
& \textbf{LD0001 (Rand)} 
& \textbf{LD0002 (Rand)} 
& \textbf{LD0004 (Rand)} 
& \textbf{LD0005 (Rand)} 
& \textbf{LD0006 (Rand)} \\
\midrule
TF-IDF Cosine
& 0.0751 & 0.0357 & 0.0648 & 0.0566 & 0.0827 & 0.0616 \\
ChrF
& 0.0834 & 0.0747 & 0.1057 & 0.1072 & 0.1553 & 0.1126 \\
Levenshtein Similarity
& 0.1122 & 0.1100 & 0.1832 & 0.1472 & 0.2187 & 0.2039 \\
BLEU
& 0.0557 & 0.0589 & 0.0913 & 0.0834 & 0.1392 & 0.0978 \\
ROUGE-L
& 0.2092 & 0.2008 & 0.3015 & 0.2546 & 0.3365 & 0.3062 \\
BERT-Score
& 0.6945 & 0.6945 & 0.7417 & 0.7168 & 0.7327 & 0.7364 \\
\bottomrule
\end{tabular}
\end{adjustbox}
\end{subtable}

\caption{Subject-wise decoding performance. (a) compares fMRI-only and fMRI+Image inputs. Bold indicates the better of the paired settings in each subject; (b) compares two baselines (Original LLaVA and Random fMRI+Image). }
\label{tab:subjectwise_two_panels}
\end{table*}

\section{Results}
\subsection{Decoding Performance}
To ensure fair comparison, all models are trained and evaluated on the same data splits. 
Following the ablation setup described in Section~\ref{sec:ablation}, we used the same data split for model training and evaluated decoding performance on the same test set.


\paragraph{Overall Metrics Comparison.}
Table~\ref{tab:subjectwise_two_panels} reports subject-wise decoding performance across six complementary evaluation metrics.
The results highlight several key findings regarding the role of neural signals in language generation.

\textbf{Semantic Guidance of fMRI Signals.}
A critical question is whether the model truly "reads" the brain or simply relies on visual priors. 
Comparing the \textit{fMRI+Image} condition in Table~\ref{tab:subjectwise_two_panels}(a) with the \textit{Random fMRI+Image} control in Table~\ref{tab:subjectwise_two_panels}(b), we observe that real fMRI signals consistently yield superior performance across all subjects. 
For instance, in subject LD0006, the BERT-Score drops from 0.7800 (Real) to 0.7364 (Random). 
This significant degradation in the absence of structured neural input provides strong evidence that the model leverages structured neural signals beyond visual priors.

\textbf{Domain-specific Prefix Tuning Effect.}
Random fMRI+Image often outperforms Original LLaVA (image-only) (e.g., 0.7417 vs 0.6945 for LD0002), suggesting that prefix tuning adapts the decoder to the dataset distribution.

\textbf{Synergistic Multimodal Decoding.}
The consistent superiority of \textit{fMRI+Image} over \textit{fMRI-only} indicates a synergistic relationship between modalities. 
While fMRI signals provide the "semantic intent" (e.g., the presence of an animal), they are inherently noisy and spatially coarse. 
The image input serves as a \textit{visual anchor}, providing structural constraints that ground the neural decoding process into specific visual details (e.g., "a yellow bird perched on a branch"). 

\textbf{Semantic vs. Lexical Alignment.}
We observe that while lexical metrics like BLEU and ROUGE-L remain relatively low, BERT-Score consistently exhibits high values (0.7--0.8). 
This discrepancy reflects the nature of brain-driven generation: the decoded text often captures the \textit{gist} of the imagined content through paraphrasing rather than exact word-for-word matching. 
The robust BERT-Score results confirm that our framework achieves stable semantic alignment, effectively bridging the gap between neural activity and natural language.


\section{Discussion}
While standard generation metrics provide an overall performance summary, they do not fully capture how well a decoded description uniquely identifies the target image among distractors.
To better analyze generation quality, we conduct a qualitative comparison of decoded descriptions, which reveals specific failure modes such as semantic drift or hallucination. We focus on representative cases.

\begin{figure*}[ht]
\centering
\includegraphics[width=1\linewidth]{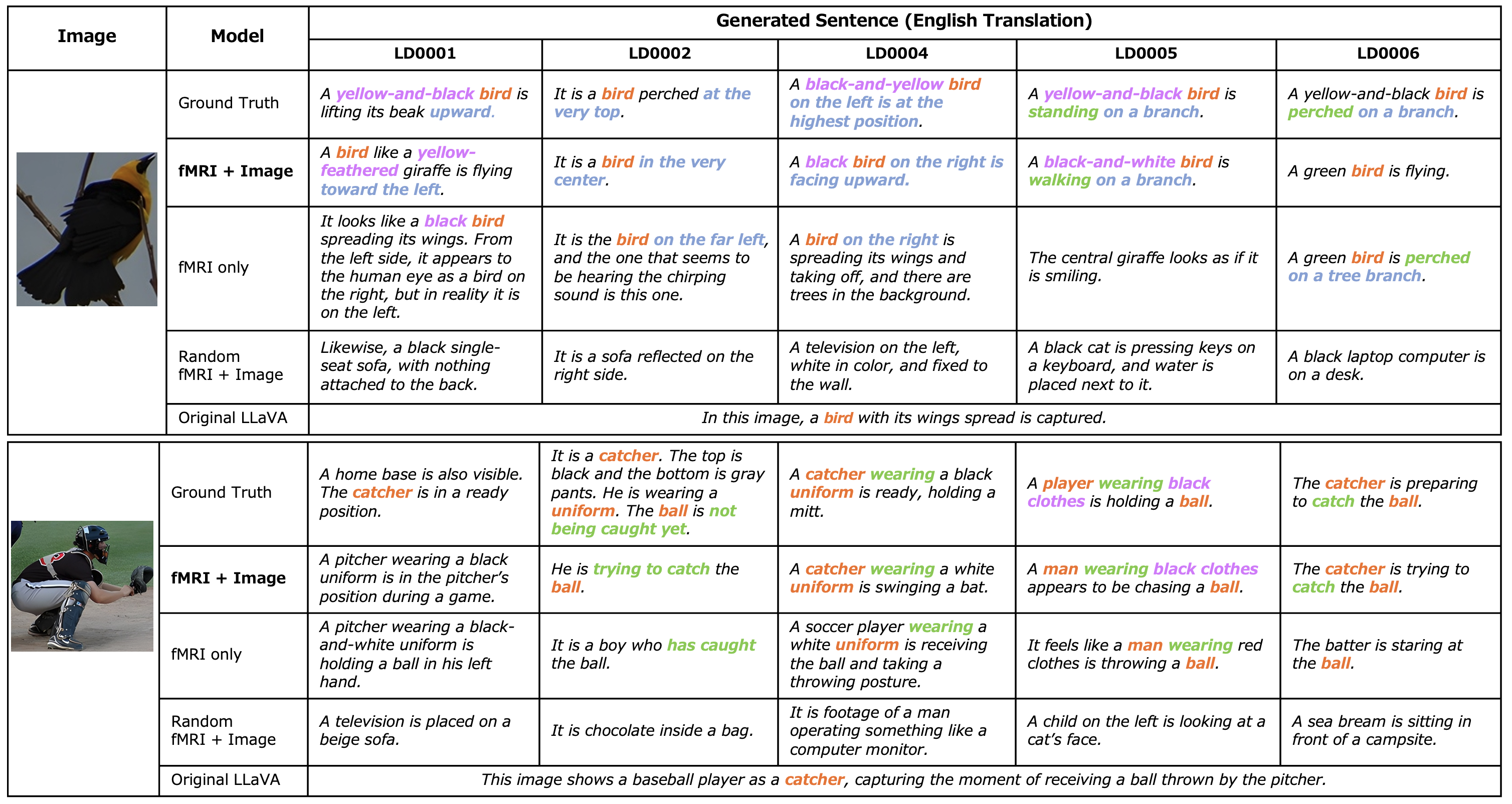}
\caption{
Decoded sentence examples (English translations) for two image–fMRI pairs.
Colors denote roles: objects (orange), actions (green), spatial (blue), color (purple).
More examples are provided in Appendix~\ref{appendix:qualitative_examples}.
}
\label{fig:example}
\end{figure*}



\begin{table}[ht]
\centering
\scriptsize
\setlength{\tabcolsep}{4pt}
\renewcommand{\arraystretch}{1.1}
\begin{adjustbox}{max width=\columnwidth,center}
\begin{tabular}{l r r r r r}
\toprule
Category & Num & LLaVA & fMRI & fMRI+Img & Rand+Img \\
\midrule
person & 20 & 0.666 & 0.703 & \textbf{0.749} & 0.707 \\
car & 5 & 0.694 & 0.699 & \textbf{0.740} & 0.702 \\
whole\_image & 45 & 0.702 & 0.741 & \textbf{0.763} & 0.723 \\
chair & 5 & 0.656 & 0.763 & \textbf{0.779} & 0.741 \\
tv & 10 & 0.711 & 0.773 & \textbf{0.788} & 0.735 \\
knife & 5 & 0.683 & 0.728 & \textbf{0.730} & 0.727 \\
broccoli & 5 & 0.662 & \textbf{0.693} & 0.687 & 0.714 \\
bus & 5 & 0.697 & \textbf{0.813} & 0.796 & 0.711 \\
\bottomrule
\end{tabular}
\end{adjustbox}
\caption{
Category-level semantic similarity (BERT-Score).
LLaVA serves as a vision-language baseline.
Bold indicates the better result between fMRI-only and fMRI+Image models.
Rand+Img is included as an ablation to verify the contribution of fMRI signals.
}
\label{tab:category_semantic_comparison}
\end{table}

\subsection{Category-Level Semantic Analysis}
\label{sec:category_analysis}

To understand which types of visual concepts benefit most from multimodal integration, we performed a fine-grained semantic analysis across different object categories. We identified the primary object category for each test sample based on its COCO metadata and computed the semantic similarity between the decoded and ground-truth sentences.

For each category $c$, the semantic score $S_c$ is defined as the average BERT-Score of the decoded sentences compared to their corresponding ground-truth descriptions. BERT-Score leverages the contextual embeddings from a pretrained multilingual BERT model to capture semantic similarity, making it particularly effective for evaluating the "gist" of brain-decoded text in Japanese.

Table~\ref{tab:category_semantic_comparison} presents category-level semantic similarity measured by BERT-Score.
Overall, the \textit{fMRI+Image} model achieves higher scores than the \textit{fMRI-only} variant for most categories, indicating that visual input generally provides complementary information to the neural signal.
The observed improvements are moderate in magnitude but consistent across multiple object categories.

For a small number of categories, the \textit{fMRI-only} model performs comparably or slightly better, suggesting that in such cases the neural signal alone already captures sufficient semantic information.
In most categories, \textit{fMRI+Image} outperforms the \textit{Rand+Img} baseline, confirming that the gains cannot be explained solely by visual priors or dataset-specific biases.

\subsection{Qualitative Analysis}
\paragraph{Semantic Contributions of fMRI Signals.}
Figure~\ref{fig:example} shows representative examples from each model. The \textit{fMRI+Image} model yields the most semantically aligned outputs, often capturing specific internal expressions imagined by the participant (e.g., “lifting its beak upward” or “yellow-feathered”). In contrast, the \textit{fMRI-only} model struggles with coherence, but occasionally produces relevant phrases (e.g., “bird” or “catcher”) despite lacking visual input. Both the \textit{original LLaVA} and \textit{random baseline} tend to generate generic scene-level descriptions (e.g., “a bird with its wings spread”), with the latter often yielding completely irrelevant outputs (e.g., “sofa” or “chocolate inside a bag”).

These observations demonstrate the distinct contributions of brain signals. The \textit{fMRI-only} model performs poorly due to the lack of visual anchors, limited training data, and high noise in fMRI signals. Interestingly, the \textit{random fMRI + image} baseline performs relatively well in terms of fluency—not because the random fMRI adds useful information, but because the decoder effectively ignores it and relies entirely on image features. This suggests that the decoder ignores the random fMRI input and relies solely on image features, highlighting the robustness of vision-language priors in the absence of informative neural signals.

In contrast, the \textit{fMRI+Image} model benefits from the integration of real neural signals, which modulate the interpretation of visual features and inject subject-specific semantic priors. This leads to more personalized and grounded descriptions that go beyond literal image captions. More qualitative examples are shown in Appendix~\ref{appendix:qualitative_examples}.

\subsection{Inter-subject Variability and Data Quality}
We observe notable variance in decoding performance across the five subjects. 
While the model architecture remains identical, the BERT-Score ranges from 0.7215 (LD0001) to 0.7800 (LD0006). 
This discrepancy can be attributed to several factors related to fMRI data quality and task compliance.

First, the \textbf{Signal-to-Noise Ratio (SNR)} of BOLD signals varies significantly between individuals due to physiological differences (e.g., cortical thickness, vascular patterns) and technical factors (e.g., head motion during scanning). 
Subjects with more stable neural responses in key semantic hubs naturally provide a cleaner target for Stage-1 alignment.

Second, the \textbf{subjective nature of inner speech} poses a unique challenge. 
Unlike passive viewing, the "silent description" task requires high levels of sustained attention and consistent mental imagery. 
Variations in how vividly a subject imagines the scene or how strictly they follow the linguistic prompts can lead to differences in the "semantic clarity" of the captured brain patterns. 
The superior performance of LD0006 suggests not only a high-quality neural signal but also a high degree of task compliance and internal consistency in their mental representations.


\subsection{Subject-agnostic Generalization}
\label{sec:cross-subject-protocol}

A key motivation of our decoupled two-stage framework is to explicitly separate subject-specific neural alignment from semantic-to-language decoding, thereby enabling subject-agnostic transfer. 
To evaluate this property, we designed a cross-subject inference protocol where the Stage-1 encoder from a \textit{source subject} is paired with a Stage-2 prefix projector trained on a different \textit{target subject}.

Importantly, Stage-1 remains strictly subject-specific, acting as a “neural normalizer” that maps individual brain activity into a CLIP-guided semantic space. 
Stage-2, in contrast, is designed to operate on subject-independent semantic representations, mapping these universal semantic embeddings into linguistic prefixes for the LLaVA decoder. 
This experiment tests whether the Stage-2 projector truly learns a generalized mapping from semantics to language, rather than overfitting to subject-specific neural distributions.
\begin{table}[ht]
\centering
\scriptsize
\begin{adjustbox}{max width=\linewidth}
\begin{tabular}{lccccc}
\toprule
S1 \textbackslash S2 & LD0001 & LD0002 & LD0004 & LD0005 & LD0006 \\
\midrule
LD0001 & \textbf{0.7215} & 0.7012 & 0.6966 & 0.7089 & 0.7129 \\
LD0002 & 0.6957 & \textbf{0.7493} & 0.7000 & 0.6982 & 0.6987 \\
LD0004 & 0.7012 & 0.7001 & \textbf{0.7159} & 0.7051 & 0.7087 \\
LD0005 & 0.7068 & 0.7049 & 0.7034 & \textbf{0.7258} & 0.7284 \\
LD0006 & 0.7196 & 0.7077 & 0.7182 & 0.7371 & \textbf{0.7683} \\
\bottomrule
\end{tabular}
\end{adjustbox}
\caption{Cross-subject decoding performance (BERT-Score). Rows represent the source subject for the Stage-1 encoder, and columns represent the target subject for the Stage-2 projector. Diagonal elements (bold) represent within-subject performance.}
\label{tab:cross_subject_matrix}
\end{table}

As shown in Table~\ref{tab:cross_subject_matrix}, 
the framework shows encouraging cross-subject transfer performance. The cross-subject scores (off-diagonal) remain competitive, exceeding 0.69 BERT-Score in all tested combinations and in some cases approaching or surpassing the within-subject performance of other subjects. For instance, the LD0006 encoder paired with the LD0005 projector achieves a BERT-Score of 0.7371, which is higher than the within-subject performance of LD0001 (0.7215).

These results support the modularity of our approach. Stage 1 aligns brain signals to a shared semantic space, while Stage 2 appears to learn a reusable semantic-to-language mapping that can transfer across subjects.

\section{Conclusion}
We show that aligning continuous fMRI signals to the embedding space of a pretrained language model yields a compact semantic representation of internally imagined speech. This representation can be used as a target for prefix tuning, enabling language generation without temporal supervision or decoder fine-tuning. The results indicate that such embedding-level alignment supports semantically meaningful decoding under limited data conditions.
\clearpage

\section*{Ethics Statement}
This study uses fMRI data collected by the authors from five human participants. During scanning, the participants viewed images selected from the COCO dataset~\cite{lin2014microsoft}, while brain activity was recorded. Written informed consent was obtained prior to the experiment. The experimental procedures were reviewed and approved by the relevant institutional ethics and safety review boards. No personally identifiable information (PII) is included in the dataset, and all analyses were conducted in accordance with established ethical guidelines for noninvasive human brain research.
\color{black}


\bibliography{colm2026_conference}
\bibliographystyle{colm2026_conference}
\clearpage
\appendix

\section{Data Collection and Validation}
\label{appendix:participant_info}
\subsection{fMRI Experimental Protocol and Participants details}
\label{app:fMRIExperimentalProtocol}
\begin{figure}[H]
    \centering
    \includegraphics[width=\linewidth]{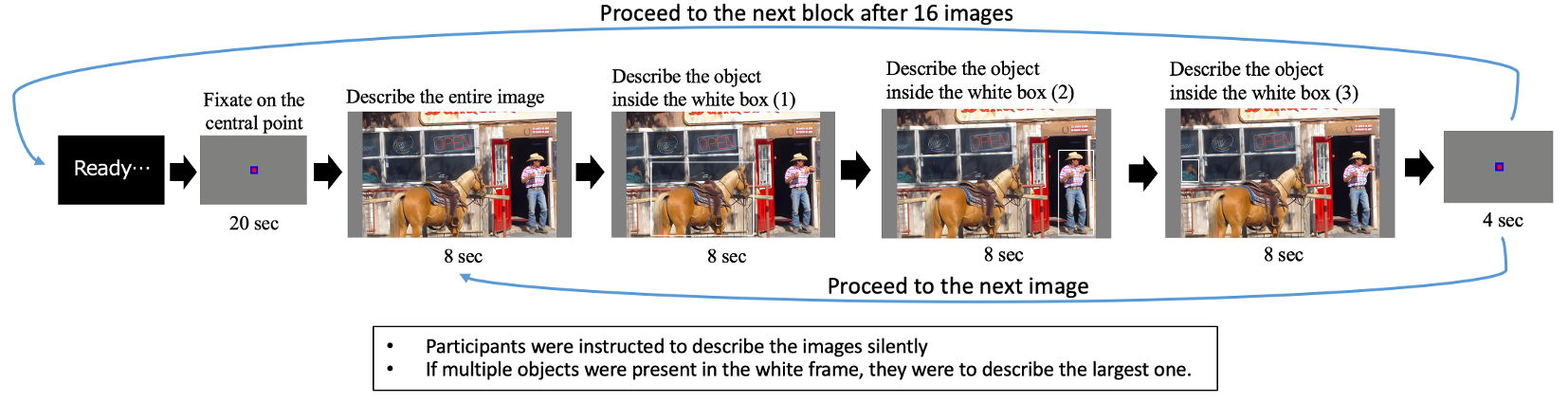}
    \caption{Experimental procedure for the inner speech task.}
    \label{fig:stimulus_example}
\end{figure}


We collected fMRI data from five healthy Japanese participants listed in Table~\ref{tab:participants} following the fMRI experimental protocol shown in Figure \ref{fig:stimulus_example}. 
The table also reports the number of cortical voxels retained after preprocessing.

\begin{table}[h!]
\centering
\small
\begin{tabular}{lccc}
\toprule
\textbf{SubjNumber} & \textbf{Age} & \textbf{Gender} & \textbf{Cortical Voxels} \\
\midrule
LD0001 & 24 & male & 277,603 \\
LD0002 & 23 & female & 250,470 \\
LD0004 & 25 & male & 284,263 \\
LD0005 & 25 & female & 240,338 \\
LD0006 & 23 & male & 299,758 \\
\bottomrule
\end{tabular}
\caption{Demographic information and cortical voxel counts of participants. 
One participant (LD0003) was excluded due to withdrawal.}
\label{tab:participants}
\end{table}

We plan to publicly release the anonymized fMRI dataset together with the accompanying code once the necessary permissions are obtained.
The data will be made available under a non-commercial, research-only license, and the code will be released under an open-source license (e.g., MIT).
This release aims to facilitate transparency, reproducibility, and further research in the field.
\color{black}

A total of 560 unique images from the COCO dataset~\cite{lin2014microsoft} were presented during the experiment. Most images were shown only once to minimize learning effects or memorization, while a small subset was repeated for validation. The image presentation was distributed over seven days. During fMRI scanning, each image was paired with four tasks: one describing the entire scene and three describing local objects highlighted with white bounding boxes. The participants performed silent inner speech (i.e., mentally composing a Japanese sentence) during each 8-second task period, without any overt articulation or lip movement.

To standardize the inner speech process and ensure task compliance, we conducted pre-scan training sessions. Participants received detailed instructions and practiced the task using example images. We confirmed that the participants could reliably generate full-sentence mental descriptions before proceeding to the actual scan.

After the scan, there was a delay of over one month before the participants were asked to write one Japanese sentence per image, recalling what were internally verbalized during the fMRI session. This design was intended to reduce any influence of short-term memory or immediate retrospective adjustment.

\subsection{Attention Monitoring and Quality Control}
To ensure alertness and attentional engagement during scanning, the participants' eye movements and blinks were continuously monitored in real time. Trials showing prolonged eyelid closure, loss of gaze fixation, or other signs of inattention were manually reviewed and excluded from the dataset. These procedures ensured that only high-quality fMRI samples with reliable internal verbalization were used in the analysis.

\subsection{Validation of Inner Speech Labels}
To validate the semantic reliability of post-scan written inner speech descriptions, we conducted a quantitative comparison against a separate set of spoken descriptions collected outside the scanner under the same image presentation protocol. For each image, the participants provided two descriptions: a written sentence reflecting inner speech, and a transcribed sentence from overt speech.

\begin{table*}[h]
\centering
\footnotesize
\renewcommand{\arraystretch}{1.2}
\begin{tabular}{|l|c|c|c|}
\hline
\textbf{Metric} & \textbf{True Pair Score} & \textbf{Random Baseline} & \textbf{Improvement} \\
\hline
Semantic Similarity (S-BERT cosine) & 0.7231 & 0.0761 & +850.8\% \\
BERT-Score (F1) & 0.7772 & 0.6600 & +17.8\% \\
\hline
\end{tabular}
\caption{Semantic similarity between inner speech and spoken description.}
\label{tab:appendix_similarity}
\end{table*}

We evaluated the semantic alignment between these two modalities using Sentence-BERT cosine similarity and BERT-Score (F1). As shown in Table~\ref{tab:appendix_similarity}, the true sentence pairs yielded a Sentence-BERT similarity of 0.7231 and a BERT-Score of 0.7772. In contrast, random sentence pairings yielded baseline scores of 0.0761 and 0.6600 respectively. This corresponds to an 850.8\% improvement in Sentence-BERT similarity and a 17.8\% gain in BERT-Score, indicating a strong alignment between the internal and spoken descriptions.

These results suggest that the post-scan written inner speech descriptions preserve substantial semantic content intended during the scan. Their consistency with overt spoken language supports their use as decoding targets for fMRI-to-language models.

\section{Explainable Variance-based Voxel Selection}
\label{appendix:ev}

To mitigate measurement noise and improve the reliability of fMRI inputs, we employ a category-based voxel selection procedure based on voxel-wise explainable variance (EV), also referred to as signal power~\cite{Sahani2002, Hsu2004, Schoppe2016}.
EV measures the proportion of response variance that is consistent across repeated presentations of the same stimulus category, thereby isolating stimulus-driven neural activity from trial-to-trial noise.

Following the variance decomposition framework introduced by Sahani and Linden~\cite{Sahani2002}, we compute the EV for each voxel $v$ within a given semantic category $c$ (e.g., COCO object categories).
For a set of $n$ repetitions belonging to category $c$, the EV is defined as:
\begin{equation}
EV_v = \frac{\mathrm{Var}(\bar{y}_v)}{\mathbb{E}[\mathrm{Var}(y_{v,i})]} - \frac{1 - EV_{\mathrm{raw}}}{n - 1},
\end{equation}
where $\mathrm{EV}_{\text{raw}}$ denotes the uncorrected explainable variance prior to bias correction,
$\bar{y}_v$ denotes the mean response of voxel $v$ across repetitions, 
and $y_{v,i}$ represents the response of voxel $v$ in the $i$-th repetition.
The second term applies a bias correction to account for finite-sample effects, 
compensating for the loss of one degree of freedom introduced by estimating the mean from the data~\cite{Schoppe2016, Reichel2025}.

Unlike conventional global EV filtering, we adopt a category-based selection strategy.
For each semantic category, we select the top $K$ voxels with the highest EV values, capturing neural populations specialized for diverse semantic contents.
A global union mask is then constructed by taking the union of category-specific voxel sets across all categories.
This union mask is applied uniformly to all samples, ensuring that the model focuses on voxels with reliable, category-relevant responses while significantly reducing input dimensionality and filtering out noise-dominated voxels.
This strategy follows established practices in fMRI encoding and decoding studies~\cite{gallantVEMtutorial}.


\section{Model Details}
\begin{figure}[h]
\centering
\includegraphics[width=0.5\linewidth]{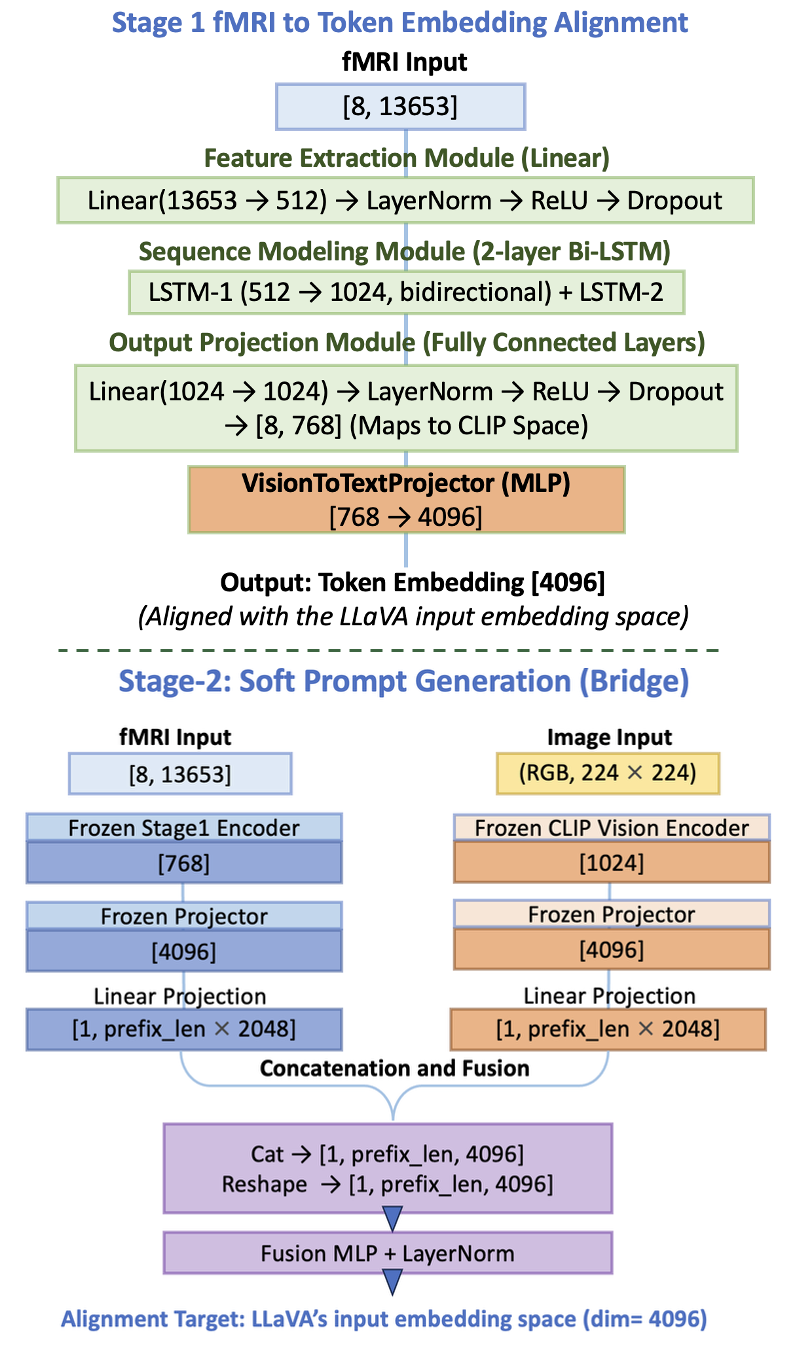} 
\caption{
Overall architecture of our two-stage brain-to-language decoding framework.}
\label{fig:method-overview}
\end{figure}

Figure~\ref{fig:method-overview} shows the overall architecture of our two-stage brain-to-language decoding framework.
Stage 1 encodes fMRI time series into a compact semantic representation aligned with the input embedding space of LLaVA.
Stage 2 constructs a multimodal soft prefix by fusing fMRI-derived semantics and image features, which is prepended to a frozen LLaVA decoder to generate text without updating decoder parameters.


\section{Token Length Statistics}
\label{sec:appendix}

\begin{figure*}[h!]
    \centering
    \includegraphics[width=1\linewidth]{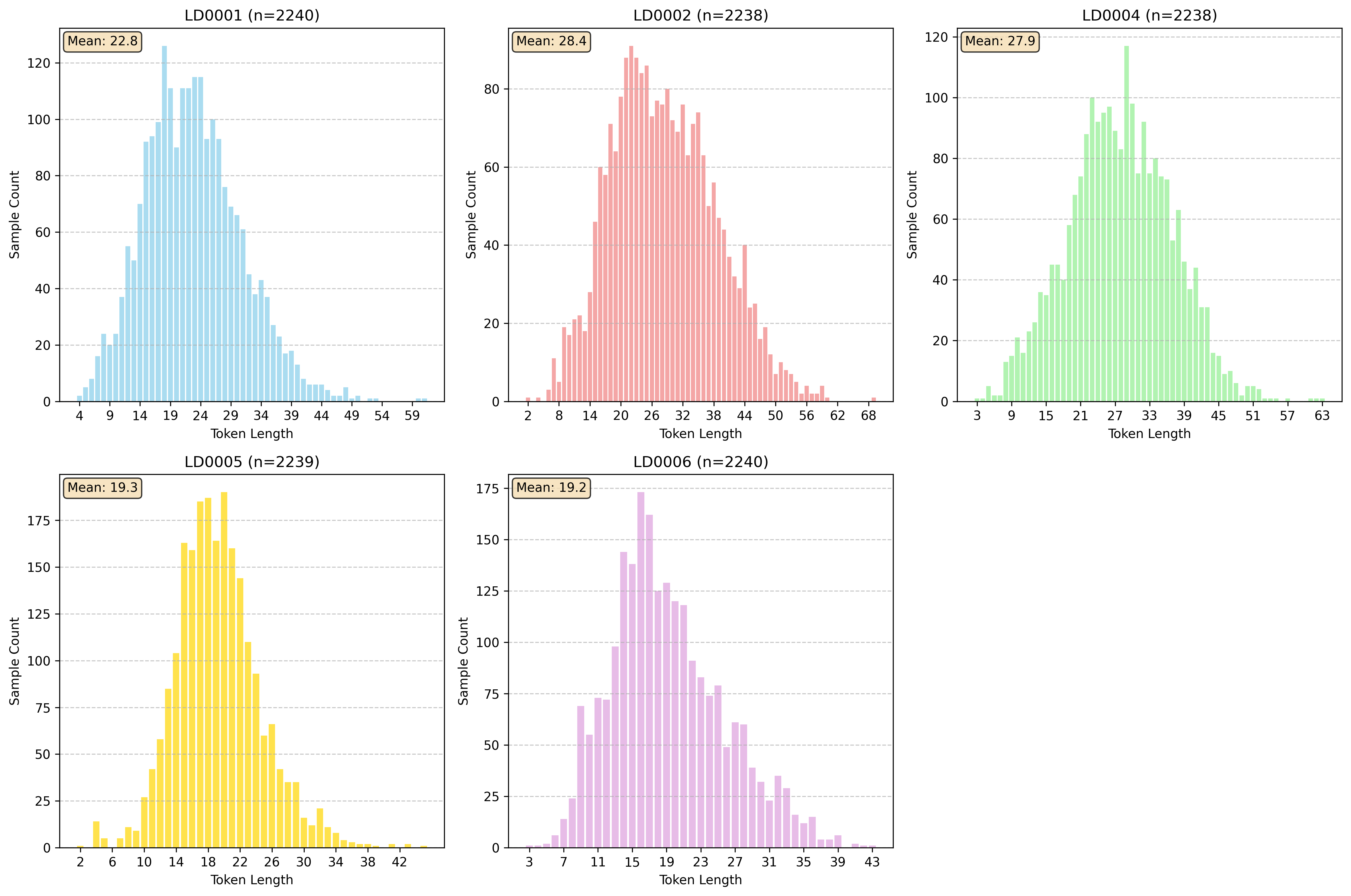}
    \caption{Histogram of token length distributions for each participant after LLaVA tokenization, showing the frequency of each token count across all 2,240 samples.}
    \label{fig:all_ld_token_distribution}
\end{figure*}

Figure~\ref{fig:all_ld_token_distribution} shows the distribution of input embedding lengths for all participants.
Each histogram represents one participant and indicates how many subword tokens were produced per sentence when their inner-speech descriptions were tokenized by the LLaVA tokenizer.
Each token corresponds to one element in the model’s input embedding sequence, rather than to a whole linguistic word.
The average sequence length ranged from about 19 to 28 tokens across participants, defining the effective input length for the fMRI-to-embedding mapping model.
\begin{table*}[h!]
    \centering
    \small
    \begin{tabular}{|c|l|c|}
        \hline
        \textbf{\#} & \multicolumn{1}{c|}{\textbf{ROI Name}} & \textbf{Primary Function} \\
        \hline
        1 & ctx\_lh\_Pole\_occipital & Visual processing \\
        2 & ctx\_rh\_Pole\_occipital & Visual processing \\
        3 & ctx\_lh\_G\_oc-temp\_lat-fusifor & Face/object recognition \\
        4 & ctx\_rh\_G\_oc-temp\_lat-fusifor & Face/object recognition \\
        5 & ctx\_lh\_G\_oc-temp\_med-Lingual & Visual processing \\
        6 & ctx\_rh\_G\_oc-temp\_med-Lingual & Visual processing \\
        7 & ctx\_lh\_G\_occipital\_middle & Visual integration \\
        8 & ctx\_rh\_G\_occipital\_middle & Visual integration \\
        9 & ctx\_lh\_G\_cuneus & Early visual processing \\
        10 & ctx\_rh\_G\_cuneus & Early visual processing \\
        11 & ctx\_lh\_G\_front\_inf-Opercular & Language production, syntactic processing \\
        12 & ctx\_lh\_G\_front\_inf-Triangul & Language production, semantic processing \\
        13 & ctx\_lh\_G\_front\_inf-Orbital & Contextual processing \\
        14 & ctx\_lh\_G\_temp\_sup-Lateral & Auditory and language processing \\
        15 & ctx\_lh\_G\_temporal\_middle & Semantic memory \\
        16 & ctx\_lh\_G\_pariet\_inf-Angular & Semantic and conceptual processing \\
        17 & ctx\_lh\_G\_pariet\_inf-Supramar & Lexical retrieval \\
        18 & ctx\_lh\_G\_oc-temp\_med-Parahip & Memory-related processing \\
        19 & ctx\_lh\_G\_temporal\_inf & Language comprehension \\
        20 & ctx\_lh\_Pole\_temporal & Language and semantic generation \\
        21 & Hippocampus & Memory formation \\
        22 & LeftHippocampus & Memory integration \\
        23 & RightHippocampus & Memory integration \\
        \hline
    \end{tabular}
    \caption{Selected 23 ROIs and their corresponding cognitive functions}
    \label{tab:roi}
\end{table*}


\section{Model Training Details}
\label{sec:training_details}

We train the Stage-1 LSTM encoder to map each fMRI sequence to a global sentence-level embedding aligned with the LLaVA input embedding space. The encoder consists of a two-layer bidirectional LSTM with 512 hidden units in each direction, followed by a projection head that outputs a 4096-dimensional representation. The outputs over the 8 fMRI steps are mean-pooled to obtain a single global embedding for each sample.

Training uses a hybrid objective composed of three core alignment losses and one auxiliary classification loss. The core alignment losses are: (1) cosine similarity loss, which encourages directional alignment between predicted and target embeddings; (2) mean squared error (MSE), which enforces element-wise similarity in magnitude and structure; and (3) an InfoNCE-based contrastive loss, which distinguishes matched fMRI--text pairs from mismatched pairs within the same batch. In addition, when valid category labels are available, we apply an auxiliary category-classification cross-entropy loss on top of the predicted global embedding. The overall training objective is:
\[
L_{\text{total}} = \lambda_{\text{nce}} L_{\text{NCE}} + \lambda_{\text{cos}} L_{\text{cos}} + \lambda_{\text{mse}} L_{\text{MSE}} + \lambda_{\text{cls}} L_{\text{cls}}.
\]

In the Stage-1 configuration used in our experiments, the loss weights are set to 1.0 for the InfoNCE term, 0.4 for cosine loss, 0.1 for MSE, and 0.2 for the auxiliary classification loss. The classification term is applied only when valid category labels are available.

Each loss term plays a complementary role. The InfoNCE objective promotes discriminative alignment by bringing each predicted embedding closer to its matched target while separating it from other samples in the batch. Cosine loss preserves semantic directionality, while MSE improves numerical fidelity by penalizing element-wise deviations. The auxiliary classification loss provides additional category-level supervision and encourages the learned representation to retain coarse semantic structure.

We optimize the model using AdamW. The batch size is 128, and the contrastive temperature is set to $\tau = 0.07$. The learning rate and the number of training epochs follow the actual Stage-1 configuration used in the experiments. Intermediate evaluations are conducted every 10 epochs.


\section{Prefix Length}
\label{appendix:prefix_length}
We conducted a small-scale ablation study using the fMRI data of participant LD0001 to examine the sensitivity of the Stage-2 model to the length of the soft prefix.
The prefix length controls the number of learnable tokens injected into the language decoder and may affect both representation capacity and computational cost.

We evaluated three prefix lengths (5, 10, and 15), while keeping all other settings fixed, including the model architecture, optimization schedule, and data splits.
Across these settings, we observed similar training dynamics and comparable validation performance, indicating that the model is relatively insensitive to moderate changes in prefix length within this range.

Given the comparable performance and to balance expressiveness with computational efficiency, we selected a prefix length of 8 for all main experiments reported in this paper.


\begin{table*}[htbp]
\centering
\scriptsize
\renewcommand{\arraystretch}{1.12}
\setlength{\tabcolsep}{4pt}
\begin{tabular}{|p{2cm}|p{3.7cm}|p{3.7cm}|p{3.7cm}|}
\hline
\textbf{Parameter} & \textbf{Stage-1 (fMRI$\rightarrow$embedding)} & \textbf{Stage-2 (fMRI+Image)} & \textbf{Stage-2 (fMRI-only)} \\
\hline
Epochs & 100 (default) & 10 (default) & 10 (same setting) \\
\hline
Learning Rate & 5e-4 (default) & 5e-5 (default) & 5e-5 (same setting) \\
\hline
Batch Size & 128 (default) & 8 (default) & 8 (same setting) \\
\hline
Weight Decay & 1e-4 (default) & 0.01 (default;\,bias/LayerNorm no decay) & 0.01 (same setting) \\
\hline
Grad Clip & 5.0 & 1.0 & 1.0 (same setting) \\
\hline
Data Split & Uses predefined split file (train/val/test) & Uses the same split as Stage-1 & Uses the same split as Stage-1 \\
\hline
Optimizer & AdamW & AdamW with decay/no-decay parameter groups & Same \\
\hline
LR Scheduler & ReduceLROnPlateau (factor=0.5, patience=5) & ReduceLROnPlateau (factor=0.5, patience=5) & Same \\
\hline
Target / Objective &
Regress a global embedding (mean-pooled over 8 steps) toward a precomputed embedding target; optional auxiliary losses (cosine/MSE/CE, etc.) &
Train a generative decoder with language modeling loss; prefix is fused from (fMRI-derived projected vector) + (image-derived vector) &
Train a generative decoder with language modeling loss; prefix uses the fMRI-derived projected vector only \\
\hline
Input Modality & fMRI only & fMRI + image & fMRI only \\
\hline
Prefix Length & --- & 8 (default) & 8 (same setting) \\
\hline
Prefix Dimension & --- & 4096 (default) & 4096 (same setting) \\
\hline
Notes (training details) &
Early stopping parameters exist (patience=10, warmup=5); augmentation flags include mixup and spatially-constrained noise (enabled by default in args) &
Encoder / vision encoder / projector are set to eval (frozen); text embedding is concatenated after prefix; prefix tokens are masked in labels &
Same training logic, without the image branch \\
\hline
\end{tabular}
\vspace{0.35em}
\caption{Training hyperparameters and configurations used in each stage (values reflect default arguments and the implemented training logic).}
\label{tab:training-protocol}
\end{table*}

\begin{table*}[t]
\centering
\small
\begin{tabular}{|l|l|c|c|c|}
\hline
\textbf{Model Variant} & \textbf{Dataset Split} 
& \textbf{Loss (Epoch 1)} 
& \textbf{Loss (Best Val Epoch)} 
& \textbf{Loss (Final Epoch)} \\
\hline
\textbf{Stage-1} (fMRI$\rightarrow$token) 
& Train      & 5.31 & N/A & 0.91 \\
& Validation & 5.01 & N/A & 3.70 \\
\hline
\textbf{Stage-2} (fMRI-only)  
& Train      & 2.20 & \textbf{1.03 (Epoch 4)} & 0.55 \\
& Validation & 1.74 & \textbf{1.49 (Epoch 4)} & 1.72 \\
\hline
\textbf{Stage-2} (fMRI+Image) 
& Train      & 3.88 & \textbf{0.27 (Epoch 9)} & 0.23 \\
& Validation & 1.73 & \textbf{0.86 (Epoch 9)} & 0.93 \\
\hline
\end{tabular}

\vspace{0.5em}
\caption{
Training and validation loss values for Subject LD0001.
Stage-1 loss measures cross-modal representation alignment and is not directly comparable to Stage-2 decoding loss.
Stage-2 results are reported at the epoch achieving the lowest validation loss.
}
\label{tab:loss_values_ld0001}
\end{table*}


\section{Selected Brain Regions of Interest (ROIs)}
\label{appendix:roi}
To focus on brain regions associated with language, memory, and visual processing, we selected 23 regions of interest (ROIs) based on prior neuroscientific literature. Table~\ref{tab:roi} summarizes their anatomical labels and primary associated cognitive functions.


\section{Training Protocol: Model Size and Computational Budget}
\label{sec:training-protocol}

We summarize the training protocol of all models in Table~\ref{tab:training-protocol}.  
Unless otherwise specified, the analyses presented in this section and all subsequent figures and tables were conducted using the fMRI data of participant LD0001.

\paragraph{Model Size.}
The Stage-1 has approximately 12 million trainable parameters and occupies about 103MB on disk.  
For Stage-2, the total model size varies with the prefix length.  
The fMRI+Image variant occupies about 15GB (including the frozen LLaVA backbone) for prefix lengths 8.  
The fMRI-only variant occupies about 14GB (including the frozen LLaVA backbone) for prefix lengths 8.  
In all configurations, the LLaVA-7B decoder is kept frozen.

\paragraph{Compute Infrastructure and Training Time.}
All models were trained on a local high-performance computing (HPC) cluster equipped with four NVIDIA RTX 6000 Ada Generation GPUs (48GB each).  
The complete training process—including both Stage-1 and Stage-2—took approximately one hour of wall-clock time using 4 GPUs in parallel, corresponding to a total of 4 GPU-hours. 


\section{Training Loss Comparison}
\label{sec:Training Loss Comparison}

\begin{table}[H]
\centering
\scriptsize
\begin{tabular}{|l|r||l|r|}
\hline
\textbf{Object Type} & \textbf{Count} & \textbf{Object Type} & \textbf{Count} \\
\hline
whole\_image & 52 & mouse & 2 \\
person & 25 & elephant & 2 \\
tv & 13 & broccoli & 2 \\
chair & 12 & bowl & 2 \\
laptop & 12 & microwave & 2 \\
car & 11 & backpack & 2 \\
oven & 8 & dining table & 2 \\
keyboard & 7 & horse & 1 \\
couch & 6 & kite & 1 \\
sink & 5 & dog & 1 \\
zebra & 5 & teddy bear & 1 \\
cup & 4 &  vase & 1 \\
bed & 4 & stop sign & 1 \\
book & 4 & toilet & 1 \\
pizza & 4 & carrot & 1 \\
cat & 4 & banana & 1 \\
bus & 3 & donut & 1 \\
orange & 3 & giraffe & 1 \\
truck & 3 & potted plant & 1 \\
airplane & 3 & bottle & 1 \\
refrigerator & 3 & knife & 1 \\
train & 2 & boat & 1 \\
bench & 2 & &  \\
\hline
\end{tabular}
\caption{Sample distribution across object types in the test set (organized in two columns for compactness).}
\label{tab:object-distribution}
\end{table}

Table~\ref{tab:loss_values_ld0001} summarizes the training dynamics of Subject LD0001 across the two training stages.
The loss values reported for Stage-1 and Stage-2 are not directly comparable, as the two stages optimize different objectives and operate on distinct learning targets.

In Stage-1 (fMRI $\rightarrow$ token embedding alignment), the model is trained to map high-dimensional and noisy fMRI signals to pretrained embeddings.
Although the training loss decreases from 5.31 to 0.91, the validation loss remains relatively high (3.70). This pattern is consistent with the difficulty of directly aligning fMRI signals with semantic representations under limited supervision and strong information compression.
Since Stage-1 is intended as a representation alignment module rather than a generative model, the validation loss is not used for model selection, and early stopping is not applied.

In Stage-2, the aligned representations produced by Stage-1 are used as soft prefixes for sequence decoding, leading to different convergence characteristics.
In the fMRI-only setting, the lowest validation loss is observed at an early epoch (Epoch~4). Beyond this point, the training loss continues to decrease while the validation loss shows a mild increase, which may indicate limited generalization when only brain signals are available.
By comparison, the fMRI+Image model reaches its minimum validation loss (0.86) at a later epoch (Epoch~9) and exhibits a smaller discrepancy between training and validation losses. This behavior is consistent with the presence of additional visual information providing supplementary semantic cues during training, which may contribute to more stable optimization and improved generalization.


\clearpage
\section{Decoding Examples}
\label{appendix:qualitative_examples}

\begin{figure}[h]
\centering
\resizebox{\textwidth}{!}{%
\begin{minipage}{\textwidth}
\centering
\includegraphics[width=\textwidth]{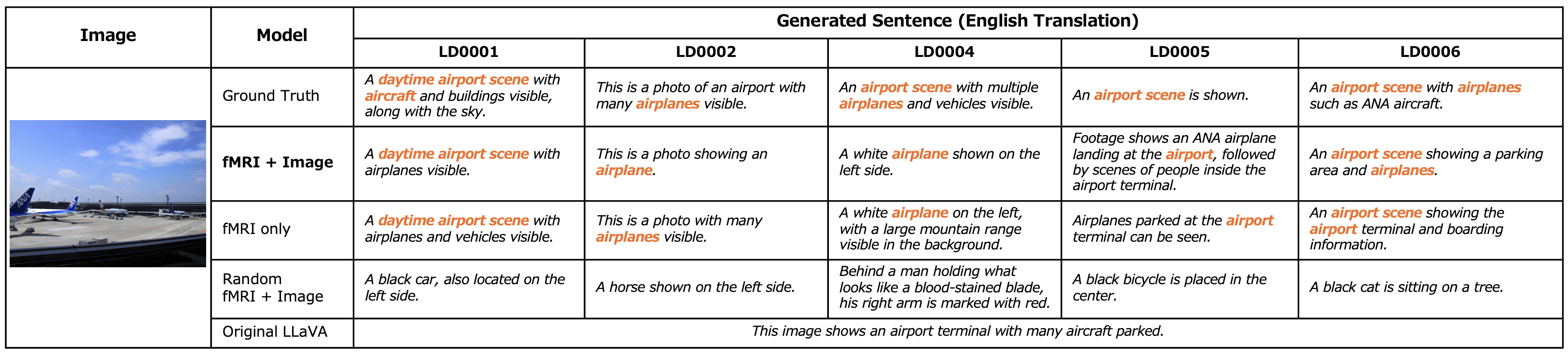}\vspace{4pt}
\includegraphics[width=\textwidth]{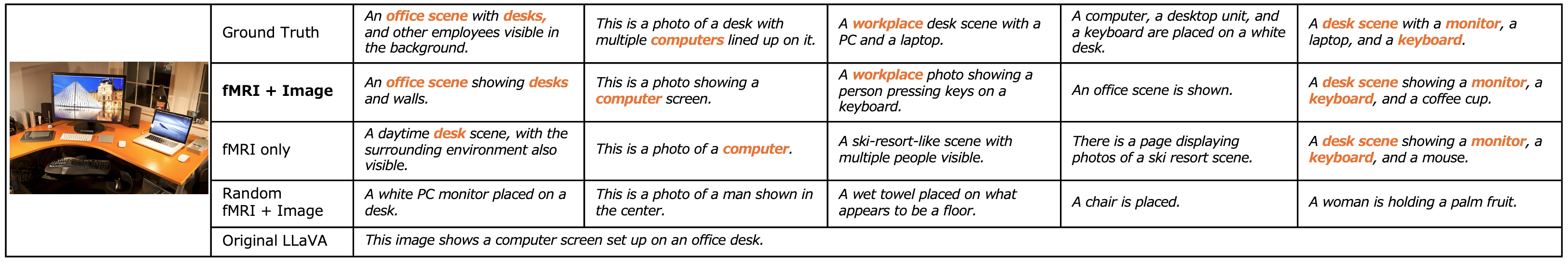}\vspace{4pt}
\includegraphics[width=\textwidth]{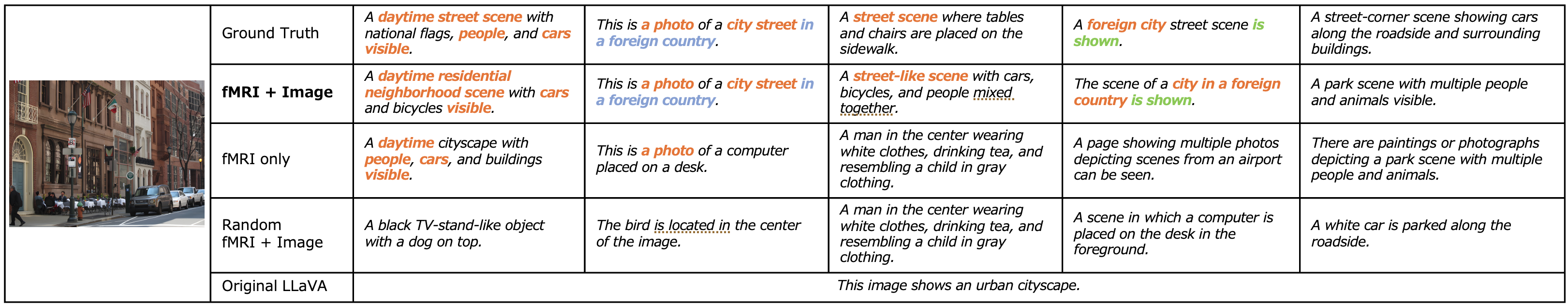}\vspace{4pt}
\includegraphics[width=\textwidth]{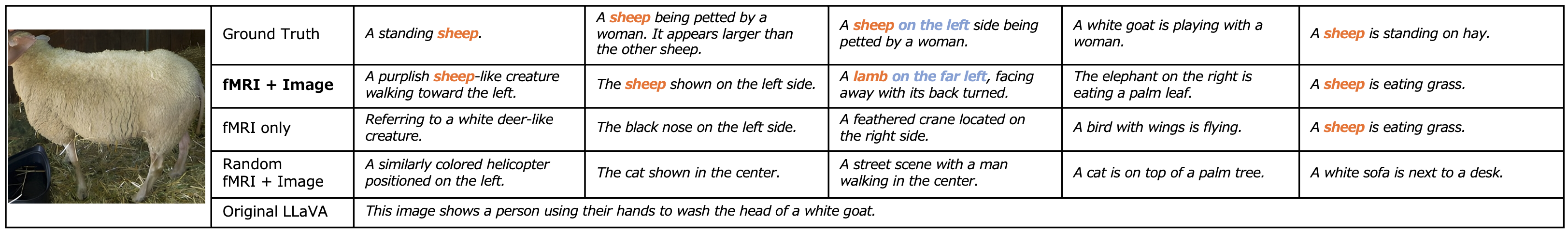}\vspace{4pt}
\includegraphics[width=\textwidth]{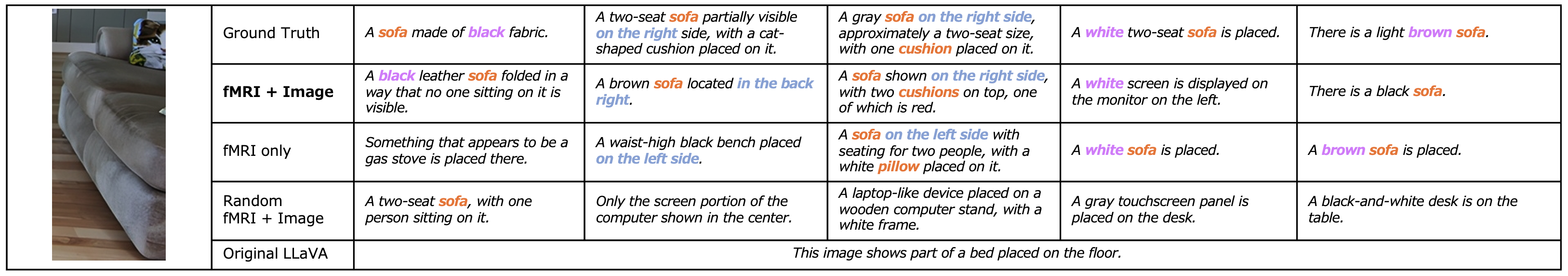}\vspace{4pt}
\includegraphics[width=\textwidth]{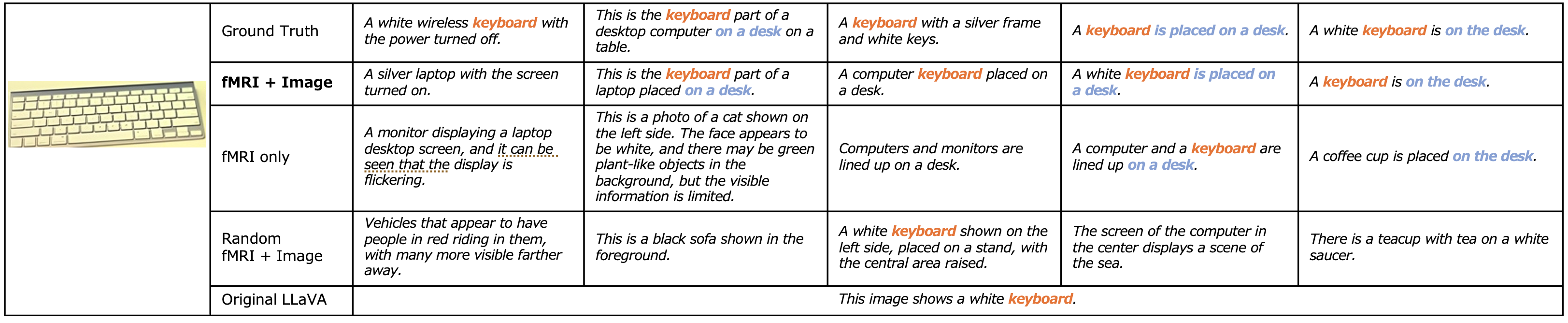}
\end{minipage}}
\caption{Qualitative decoding examples.}
\end{figure}

%






\end{document}